%% file: main.tex
\pdfoutput=1

\documentclass[11pt]{article}

\usepackage[]{acl}

\usepackage{times}
\usepackage{latexsym}

\usepackage[T1]{fontenc}

\usepackage[utf8]{inputenc}

\usepackage{microtype}
\usepackage{algorithm}
\usepackage{algorithmic}
\usepackage[utf8]{inputenc} 
\usepackage[T1]{fontenc}    
\usepackage{url}            
\usepackage{booktabs}       
\usepackage{amsfonts}       
\usepackage{nicefrac}       
\usepackage{microtype}      
\usepackage{xcolor}         
\usepackage{enumitem}
\usepackage{multirow}
\usepackage{graphicx}
\usepackage{amsmath}

\input{macros}

\title{Dr.ICL: Demonstration-Retrieved In-context Learning}

\author{
    Man Luo$^1$ \quad Xin Xu$^2$ \quad Zhuyun Dai$^2$ \quad \textbf{Panupong Pasupat}$^2$ \\ \quad \textbf{Mehran Kazemi}$^2$ \quad \textbf{Chitta Baral}$^1$ \quad \textbf{ Vaiva Imbrasaite}$^2$ \quad \textbf{Vincent Y Zhao}$^2$ \\
    \textsuperscript{1} Arizona State University \quad \textsuperscript{2} Google Research\\
    \texttt{\{mluo26, chitta\}@asu.edu} \\
    \texttt{\{xxujasmine, 
zhuyundai, 
ppasupat, mehrankazemi, vimbrasaite, 
vzhao\}@google.com}
}

\begin{document}
\maketitle

\input{abstract}

\input{introduction}


\input{related_work} 
\input{approach}
\input{experiments}
\input{analysis}

\input{conclusion}

\bibliography{custom}

\appendix
\section{Examples of Retrieved Demonstrations}
\input{tables/examples}



\end{document}

%% file: macros.tex
\newcommand{\dricl}{Dr.\ ICL}

%% file: abstract.tex
\begin{abstract}
    In-context learning (ICL), teaching a large language model (LLM) to perform a task with few-shot demonstrations rather than adjusting the model parameters, has emerged as a strong paradigm for using LLMs. While early studies primarily used a fixed or random set of demonstrations for all test queries, recent research suggests that retrieving semantically similar demonstrations to the input from a pool of available demonstrations results in better performance. This work expands the applicability of retrieval-based ICL approaches by demonstrating that even simple word-overlap similarity measures such as BM25 outperform randomly selected demonstrations. Furthermore, we extend the success of retrieval-based ICL to instruction-finetuned LLMs as well as Chain-of-Thought (CoT) prompting. For instruction-finetuned LLMs, we find that although a model has already seen the training data at training time, retrieving demonstrations from the training data at test time yields better results compared to using no demonstrations or random demonstrations.
    Last but not least, we train a task-specific demonstration retriever that outperforms off-the-shelf retrievers.
\end{abstract}

%% file: introduction.tex
\section{Introduction}

Language models are now the foundation models for many natural language processing tasks across a wide range of domains~\citep{bommasani2021opportunities}. 
One of the most exciting emergent abilities~\citep{weiemergent} of large language models (LLMs) is in-context learning (ICL)~\citep{brown2020language,mishra2022cross}. With ICL, instructions and a few demonstrative examples are augmented to the inputs to LLMs, allowing them to perform well on new tasks without the need for fine-tuning.

\begin{figure}[t]
\centering
\includegraphics[width=\linewidth]{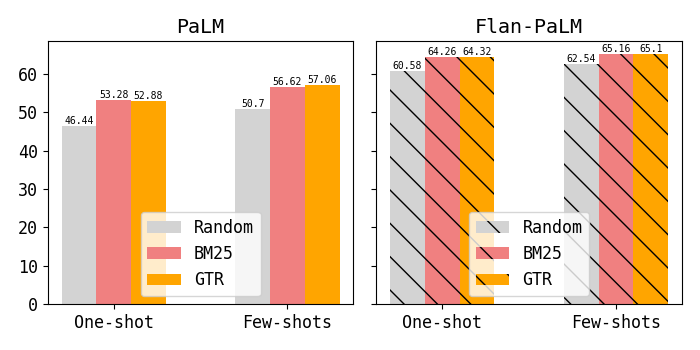}
\caption{The average performance of PaLM and Flan-PaLM on five datasets, with one and few-shot ICL. Retrieved demonstrations given by either BM25 or GTR yield better performance than random demonstrations.}
\label{fig:flanpalm_avg}
\end{figure}


Typically, ICL approaches utilize random or hand-crafted demonstrations that are applied across various queries. This may, however, not always be optimal. Recent research has revealed that using demonstrations semantically similar to the input query can enhance performance~\cite{liu2022makes}. 
In this work, we investigate two off-the-shelf retrievers,  BM25~\cite{robertson2009probabilistic} and GTR~\cite{ni2021large}, where BM25 is a sparse retriever that finds demonstrations with the highest (weighted) word overlap with the query, while GTR is a dense retriever that seeks demonstrations semantically closest to the query.
Then, we utilize them to obtain query-specific demonstrations, and study demonstration-retrieved ICL (\dricl{}) with a general and an instruction-finetuned LLM. 
Beyond previous work, several interesting findings are discovered through our experiments as shown in Figure \ref{fig:flanpalm_avg}.
Firstly, we establish that both BM25 and GTR can find more effective demonstrations than random demonstrations in both one-shot and few-shot ICL settings.
Such off-the-shelf retrievers make \dricl{} an appealing paradigm for real-world applications.  
Secondly, our results with an instruction-finetuned LLM, i.e., Flan-PaLM~\citep{chung2022scaling}, indicate that training data can be useful not only for training models but for accompanying a retriever for testing, suggesting a more efficient way to utilizing training data which are expensive to collect.
Lastly, by combining with an advanced prompting technique, Chain-of-Thought (CoT)~\cite{wang2022towards}, demonstration-retrieved proves to be more effective than relying solely on CoT. 
This suggests that \dricl{} can boost the performance of powerful prompt engineering techniques. 

Next, we aim to go beyond off-the-shelf retrievers which are often geared towards question answering or information retrieval tasks thus the retrieved demonstrations might capture query-specific knowledge required to answer the query. 
However, the retrieved demonstrations given by the off-the-shelf retrievers might not represent the nature of the task and how the task should be solved in general. 
Consider, for example, the query ``In a barn are chickens and rabbits with 35 heads and 94 legs total. How many chickens and rabbits are there?''. Off-the-shelf retrievers may mostly provide information about the animals in the question and their properties such as number of heads and legs (i.e. query-specific knowledge), but may not provide enough similar linear algebra questions (i.e. information about the nature of the task).

Therefore, we develop a demonstration retriever that is tailored to retrieving representative demonstrations.
Figure~\ref{fig:rap} showcases the process of training the demonstration retriever: we first create a demonstration retrieval training set using signals from a language model. Concretely, we use an off-the-shelf retriever to find demonstration candidates for a given input question, prepend them to the question, and then obtain probabilities from the language model to re-rank the candidates. We then use the top-$n$ and bottom-$n$ candidates as positive and hard-negative examples, respectively, to construct a training set and train the retriever to identify the best demonstration example for a given query.
Experimental results show that the demonstration retriever outperforms off-the-shelf retrievers, with more noticeable improvement in one-shot ICL. This encouraging result indicates that the trained retriever could offer an effective substitute for off-the-shelf models.

%% file: related_work.tex
\section{Related Work}

\input{tables/related_work}

\subsection{Few-shot In-context Learning}

Few-shot in-context learning (ICL) is a technique that allows language models, such as GPT-3~\cite{brown2020language} and PaLM~\cite{chowdhery2022palm}, to generalize to new tasks based on a small number of examples. 
ICL offers several advantages over the traditional training approach of language models, which involves pre-training followed by fine-tuning. One key benefit is that fine-tuning may not always be feasible due to restricted access to the LLM or inadequate computational resources~\citep{brown2020language}.
Additionally, ICL avoids the issues commonly associated with fine-tuning, such as overfitting~or~shocks~\cite{ying2019overview,kazemi2023understanding}, as it does not modify the model's parameters, allowing it to remain general.
However, the effectiveness of ICL varies depending on various factors, such as the order of the demonstrations~\cite{kumar2021reordering}, the distribution of the demonstrations~\cite{min2022rethinking}, and the complexity and quality of the prompts themselves~\cite{zhao2021calibrate,arora2022ask}. Some research has shown that lower perplexity prompts~\cite{gonen2022demystifying} and open-ended question-answer formats~\cite{arora2022ask} tend to lead to better performance, while others have found that intermediate reasoning steps~\cite{wei2022chain} and higher complexity prompts~\cite{fu2022complexity} can also improve results on certain tasks~\cite{suzgun2022challenging,wang2022towards}.
In an effort to understand how ICT works, studies have suggested that ICL may involve implicit Bayesian inference~\cite{xie2021explanation} and a symbiotic relationship between text and patterns~\cite{madaan2022text}, and can behave similarly to explicit fine-tuning~\cite{dai2022can}.
Our work focus on the effect of demonstrations for ICL with large language models. 

\subsection{Retrieval Augmented Demonstrations}

As summarized in Table~\ref{tab:related_work},
several previous works have explored retrieval techniques for identifying more informative demonstrations to boost in-context learning.
KATE~\cite{liu2022makes} discovers that semantically closer demonstrations outperform random ones for GPT-3 in-context learning. They employ language models trained on tasks like natural language inference and sentence textual similarity as semantic representations and utilize the kNN algorithm to search for demonstrations.
EPR~\cite{rubin2021learning} develops a retriever based on language model signals to find superior demonstrations compared to off-the-shelf retrievers. Instead of using a separate retriever for each task, UPRISE~\citet{cheng2023uprise} merges multiple training datasets into a retrieval corpus and trains a universal retriever for cross-domain tasks.
PARC~\cite{Nie2022CrossLingualRA} employs a multilingual retrieval strategy to find demonstrations from high-resource tasks, thereby enhancing the performance of low-resource domain tasks. CEIL~\cite{ye2023compositional}, instead of retrieving few-shot demonstrations independently, introduces an iterative retrieval method to identify both diverse and similar few-shot examples.
While the aforementioned methods retrieve demonstrations from training data, \citet{madaan2022memory,dalvi2022towards} incorporate human feedback to create demonstrations and maintain a dynamic retrieval corpus. Z-ICL~\cite{lyu2022z} generates pseudo demonstrations to enhance zero-shot in-context performance.
In contrast to the methods that retrieve explicit demonstrations, RETROPROMPT~\cite{chen2022decoupling} transforms explicit demonstrations into implicit neural demonstrations represented by vectors. Rather than using a retriever, ~\citet{ram2023context} applies a cross-attention reranker to re-rank documents retrieved by BM25.





%% file: tables/related_work.tex
\begin{table*}[t]
    \centering
    \small
    \resizebox{0.95\linewidth}{!}{
    \setlength{\tabcolsep}{5pt}
    \begin{tabular}{@{}p{0.1\linewidth}p{0.1\linewidth}p{0.2\linewidth}p{0.2\linewidth}p{0.2\linewidth}p{0.2\linewidth}p{0.1\linewidth}@{}}
        \toprule
        \textbf{Paper} & \textbf{LLMs}  & \textbf{Retrieval Method}  & \textbf{Retrieval Corpus} & \textbf{Evaluation Tasks}& \textbf{ \# of Shots in Prompts} & CoT\\
        \toprule
    KATE~\citeyear{liu2022makes} & GPT-3 & RoBERTa+kNN & In-Domain TD & SA, T2T & Few-shots & No \\
    \midrule
    EPR~\citeyear{rubin2021learning} & GPT-J, GPT-Neo,  CODEX, GTP-3 & SBERT, BM25, FT Retriever & In-Domain TD & SRM& Few-shots& No \\
    \midrule
    CEIL~\citeyear{ye2023compositional}& GPT-Neo, GPT2-XL, CodeX & BM25, BERT, DPR, FT Retriever & In-Domain TD & SA, PD, NLI, CSR, QA, codeG, and SP & Few shots & No\\
    \midrule
    UPRISE~\citeyear{cheng2023uprise} & GPT-Neo, BLOOM, OPT, GPT-3 & FT Retriever & Cross Tasks TD & RC, QA, NLI, SA, CSR, CR, PD & Few shots& No\\
    \midrule
    Ours & PaLM, Flan-PaLM & BM25, GTR, FT Retriever &  In-Domain TD & QA, NLI, MathR, BC & One-shot, Few-shots & Yes \\
    \bottomrule
    \end{tabular}
    }
    \caption{Comparison with Related Work. TD: training data, QA: question answering, RC: reading comprehension, NLI: natural language inference, SA: sentiment analysis, CSR: commonsense reasoning, CR: Coreference Resolution, MathR: mathmatical reasoning, PD: paraphrase detection, SP:semantic parsing, CodeG: code generation, SRM: Sentence representation mapping, T2T: Table to Text generation, Question Answering, 
    }
    \label{tab:related_work} 
\end{table*}

%% file: approach.tex
\section{Demonstration-Retrieved In-Context Learning (\dricl{})}
\label{sec:approach}

\begin{figure*}[t]
\centering
\includegraphics[width=\linewidth]{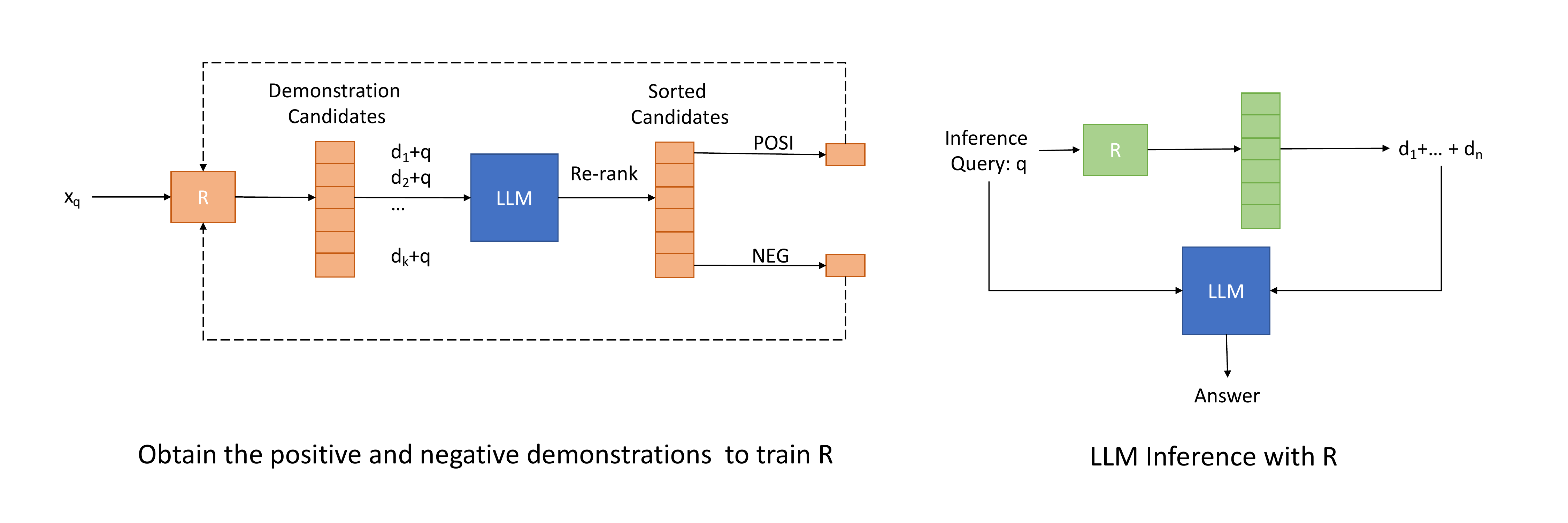}
\caption{Pipeline for training demonstration retriever and inference (R for a neural retriever). Figure on the left shows the procedure of obtaining data to train a demonstration retriever: an off-the-shelf retriever takes an input query $x_{q}$ and retrieves top-$k$ (e.g., 100) demonstrations candidates from the training corpus. Then an LLM is used to output the score of the ground truth of $y_{q}$ with each retrieved demonstration and $x_{q}$. Figure on the right shows the inference pipeline for in-context learning with the trained demonstration retriever.}
\label{fig:rap}
\end{figure*}

We start by describing ICL for general tasks (including classification or generation tasks). 
For a task $T$, given an input text $x_{q}$, an LLM is used to predict the answer $y_{q}$ conditioned on a set of \emph{demonstrations} of the task, $Demo = \{d_1, d_2,\cdots, d_n\}$, where $d_{i} = (x_i, y_i)$ is a pair of input and ground truth answer.
Typically, $d_{i}$ is linearized as a string (e.g., ``\texttt{question:~$x_{i}$~\textbackslash{}n~answer:~${y_{i}}$}'') and then provided to the LM. Recently, the Chain-of-thoughts prompting technique~\cite{wei2022chain} has demonstrated its effectiveness in handling complex reasoning tasks. The primary concept involves including intermediate reasoning steps for each demonstration, so it consists of not only the input and ground truth answer but also the step-by-step reasoning process.


There are multiple strategies for choosing the set of demonstrations.
For instance, one could randomly or manually select a fixed set $Demo$ to be applied to all queries of task $T$.
Alternatively, a retriever can be used to find query-specific demonstrations from the training set $D_{train}$:
\begin{equation}
    Demo_{x_{q}} = Retriever(x_{q}, D_{train},n),
\end{equation}
where $Demo_{x_{q}}$ are the top-$n$ demonstrations that the retriever considers most suitable for the input $x_{q}$.
In this work, we consider two off-the-shelf retrievers, BM25 and GTR (Section~\ref{sec:off-the-shelf}), and then propose a method to train a retriever tailored to the target task $T$ (Section~\ref{sec:training}).

\subsection{Off-the-shelf Retrievers}
\label{sec:off-the-shelf}

BM25~\cite{robertson2009probabilistic} is a bag-of-words model that calculates relevance scores using term frequency, inverse document frequency, and document length normalization. It has proven effective and efficient, making it easily deployable in large-scale, real-world applications. However, BM25 heavily relies on keyword matching and lacks context understanding, which may result in less accurate outcomes. In contrast, GTR~\cite{ni2021large} is a dual-encoder neural retriever (based on T5) trained on the MS Marco dataset~\cite{nguyen2016ms}. GTR excels in semantic and context comprehension and is easily transferable to downstream tasks or specific domains. However, it has lower memory and computational efficiency, and lacks interpretability.

\subsection{Demonstration Retriever Training}
\label{sec:training}
Demonstration retrieval aims to find the most representative demonstrations for each input query.
Ideally, the demonstrations should capture both (a) the query-specific knowledge required to answer the query, and (b) the nature of the task and how the task should be solved in general. 

Off-the-shelf retrievers such as BM25 and GTR were designed for information retrieval and question answering. As such, they mostly retrieve demonstrations of type (a) but not (b).
To fill this gap, we propose to train a demonstration retriever by leveraging the feedback from a language model. As demonstrated in Figure~\ref{fig:rap}, the process involves two steps: obtaining the training data and training a retriever on the data. 

\paragraph{Obtain the Training data}
We want to teach the retriever model to locate examples that lead to the most accurate predictions.
We propose to mine a set of demonstrations for each input query $x_q$ in the training data as follows.
First, given a question-answer pair $(x_{q}, y_{q}) \in D_{train}$, we use an off-the-shelf retriever to find a demonstration candidate set $D$ for $x_{q}$, where $x_{q}$ is exclusive from $D$. Second, we test each demonstration $d \in D$ on how much it helps on the target task. The LM probability $p_\text{LM}(y_q \mid d, x_q)$ of the gold answer $y_q$ is used as the score for the demonstration.
Finally, we keep the top-$n$ demonstration as the positive demonstrations, and the bottom-$n$ as the hard negative demonstrations.

\paragraph{Training Procedure}
Our retriever is a dual encoder, which defines the score of any query-document pair $(q, d)$ as $s(q,d) = v_q^\top v_d$, where $v_q$ and $v_d$ are the embeddings of $q$ and $d$.
We initialize our retriever with GTR, and then fine-tune it on the training data via contrastive loss with both in-batch and hard negatives:
\begin{equation}
    \mathcal{L}_{con} = -\log \frac{e^{s(q, d^+)}}{e^{s(q, d^+)} + \sum_j e^{s(q, d_j^-)}},
    \label{eq:dpr_loss1}
\end{equation}
where $d^+$ and $d_j^-$ are the positive and negative demonstrations. The negative demonstrations include the positive demonstrations for the other input queries in the same batch and 1 randomly-chosen hard negative demonstration.

%% file: experiments.tex
\begin{figure*}[t]
\centering
\includegraphics[width=\linewidth]{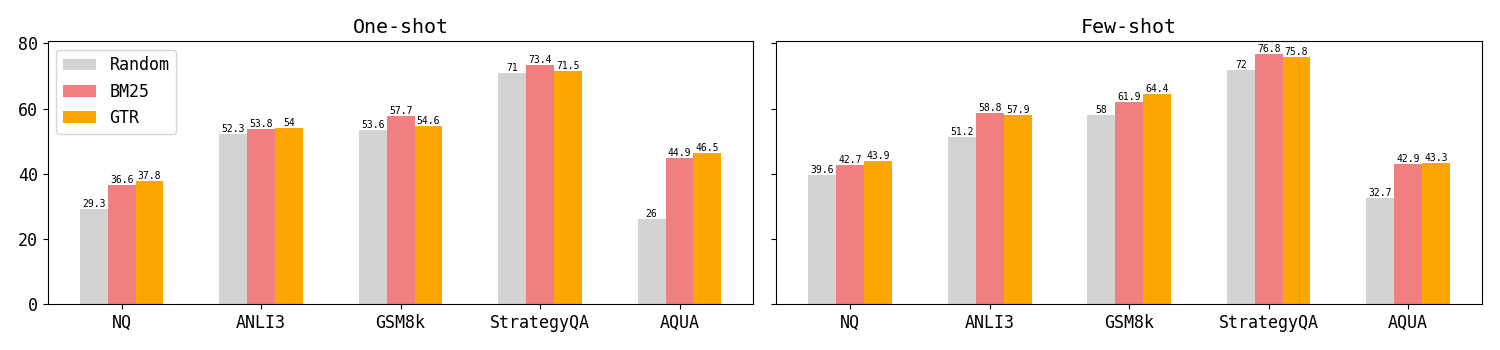}
\caption{PaLM: One-shot and few-shot inference with three types of demonstrations, random, BM25, and GTR.
}

\label{fig:palm_offshelf}
\end{figure*}

\begin{figure*}[t]
\centering
\includegraphics[width=\linewidth]{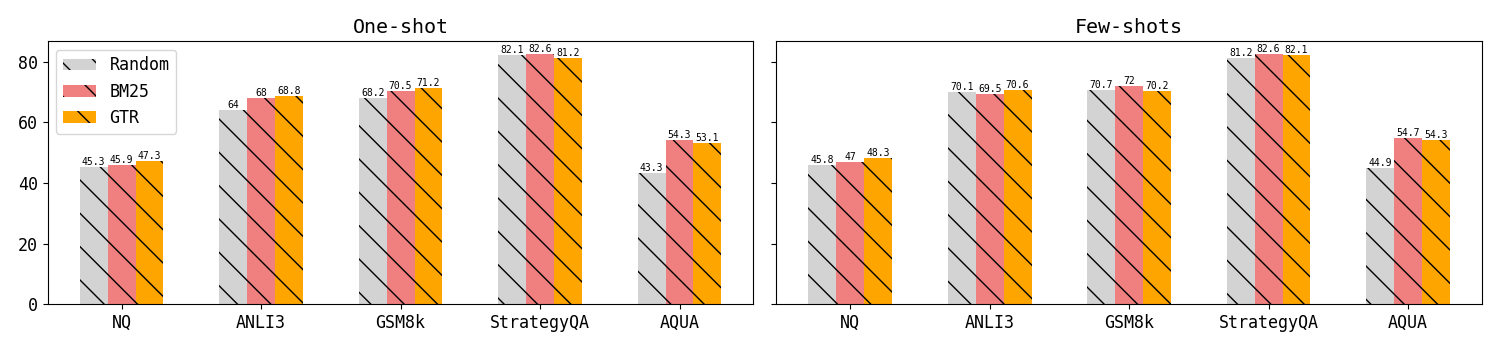}
\caption{Flan-PaLM: One-shot and few-shot inference with three types of demonstrations, random, BM25, and GTR.}
\label{fig:flanpalm_offshelf}
\end{figure*}

\section{Experiments}

\paragraph{Datasets and Evaluation Metrics}
We study various tasks across 5 datasets: free-form question answering (NQ), natural language inference (ANLI-r3), mathematical reasoning (GSM8k and AQuA) and boolean question answering (StrategyQA).
For the last three datasets, we apply CoT.
All the tasks are evaluated by exact matching accuracy.

\paragraph{Language Models}

PaLM-540B~\cite{chowdhery2022palm} and Flan-PaLM (540B)~\cite{chung2022scaling} are used as the primary LLMs. Both models have the same architecture, but Flan-PaLM has been further trained on thousands of tasks for instruction learning (including all the five datasets we studied in this paper) and shows superior generalization performance compared to PaLM.
At inference time, we use the temperature of 0.0 and maximum decoding length 10 for tasks without CoT and 256 for tasks involving CoT. 

\paragraph{Retrievers}
As explained in \S\ref{sec:approach},
we explore using BM25 and GTR as off-the-shelf retrievers, as well as training our own retriever for each task.

For BM25, we use uncased BERT wordpiece tokenization and parameters $(k_1, b) = (1.5, 0.75)$.
For GTR, we use the pretrained GTR-Base model.

When mining data for training our retriever, we use the pretrained GTR to retrieve 100 demonstrations candidates, and then use PaLM-62B to score each candidate. (We used the smaller PaLM-62B instead of 540B for efficiency.)
Then we select the top-5 reranked demonstrations as the positive candidates to fine-tune GTR.

\paragraph{Retrieval Corpus}
We create a separate retrieval corpus for each task using the associated training data. 
For tasks with CoT, each entry in the corpus is composed of the question, the CoT, and the answer, while for other tasks are without the CoT.

\subsection{Results}

\paragraph{Off-the-shelf-retriever performance}
Figures~\ref{fig:palm_offshelf} and~\ref{fig:flanpalm_offshelf} show the performance of PaLM and Flan-PaLM under one-shot and few-shot ICL settings, with and without retrievers. 
We make the following observations. 

\textit{Observation 1: Off-the-shelf retrievers are capable of finding more effective demonstrations than random ones.} 
Figure \ref{fig:palm_offshelf} shows that the demonstrations retrieved by BM25 or GTR are better than random ones under both one-shot and few-shot scenarios for the PaLM model.
It is worth mentioning that BM25 is more efficient in terms of indexing memory and retrieval latency compared to semantic retrievers like GTR or other sentence encoders~\cite{liu2022makes}, which makes it easier to deploy.

\textit{Observation 2: \dricl{} improves instruction-finetuned LLM.}
Previous research has primarily focused on investigating demonstration retrieved ICL with general LLMs (such as GPT-3) rather than instruction-finetuned LLMs, possibly because they did not consider reusing the training data. In our study, we examine \dricl{} with Flan-PaLM, an instruction-finetuned LLM, and present the results in Figure \ref{fig:flanpalm_offshelf}. Overall, the retrieved demonstrations outperform no demonstrations or random demonstrations. This implies that the training data should be reused during inference as they can be retrieved and enhance the performance, even if the model has already seen such data. We conjecture that the retrieved demonstrations may enhance knowledge localization for ICL, which could explain the observed improvement.

\textit{Observation 3: \dricl{} can further improve advanced prompting technique, Chain-of-Thought.}
In our experiments on GSM8k, StrategyQA, and AQuA, using \dricl{} in conjunction with CoT results in improved performance under both one-shot and few-shot ICL scenarios. This finding suggests that \dricl{} has the potential to enhance the performance of powerful prompting techniques.  



\medskip
\noindent The observations above hold significant values for real-world applications.
Incorporating ICL with a simple BM25 demonstration retriever, which is highly scalable in terms of latency and indexing memory, is proven to improve the performance of the LLM, including when instruction finetuning or Chain-of-Thought were used.
Examples of retrieved demonstrations given by the off-the-shelf retrievers are given in the Table~\ref{tab:example} in Appendix.

\paragraph{Trained Demonstration Retriever Performance} 
\input{tables/palm_train_retriever}
We experiment our trained demonstration retriever with PaLM.
Table \ref{tab:palm_trained_retriever} displays both one-shot and few-shot performance and show that the demonstration retriever is better than off-the-shelf GTR in almost all cases, leading to a better overall performance.
Notably, the improvements were most significant in the one-shot ICL scenario, which requires less inference latency and computing resources than few-shot ICL. These promising results suggest that the trained retriever could provide an effective alternative to off-the-shelf models. 

%% file: tables/palm_train_retriever.tex


\begin{table}[t]
    \centering
    \small
    \resizebox{0.95\linewidth}{!}{
    \setlength{\tabcolsep}{5pt}
    \begin{tabular}{@{}cccc@{}}
        \toprule
        \textbf{Task}  & \textbf{Method} & \textbf{One Shot}  & \textbf{Few Shots} \\
        \toprule
    \multirow{2}{*}{{NQ }} 

    & GTR       & 37.8  & 43.9 \\
    & Demo-GTR(our)  & \textbf{39.2(+1.4)}  & 43.9 \\
    \midrule
    \multirow{2}{*}{{ANLI (r3)}}
     & GTR       & 54.0  & 59.0  \\
    & Demo-GTR(our)  & \textbf{54.8(+0.8)}  & 59.0 \\
    \midrule
    \multirow{2}{*}{{GSM8k}}
    & GTR       & 57.7  & 61.0\\
    & Demo-GTR(our)  & \textbf{59.3(+1.6)}  & \textbf{61.5(+0.5)}\\
    \midrule
    \multirow{2}{*}{{Avg.}}
    & GTR  & 49.8 & 54.6    \\
    & Demo-GTR(our) & \textbf{51.1(+1.3)} &\textbf{ 54.8(+0.2)} \\
    \bottomrule
    \end{tabular}
    }
    \caption{ Performance of PaLM using GTR and Demo-GTR retrieved demonstrations. Demo-GTR consistently achieves better performance than GTR in one-shot case.
    }
    \label{tab:palm_trained_retriever} 
\end{table}
        

%% file: analysis.tex

\section{Analysis}

To rule out the chance that retrieved demonstrations are more advantageous than random ones simply because in the benchmark datasets the former's answers are identical to the correct ones, we assess the overlap percentage between the demonstration responses and the target.
In the few-shot scenario, we aggregate the answers from the demonstrations via majority voting.
From Table \ref{tab:analysis_fewshot}, it is evident that for the first forth datasets, the overlap ratio is roughly equal to or less than the uniform distribution, suggesting that the benefits of the retrieved demonstrations are not due to label identification.
In the case of the NQ, we notice a considerable overlap between demonstration answers and the ground truth. We then randomly select 100 instances out of the 433 overlapped cases from GTR-retrieved demonstrations (one-shot) and manually examine them. 
We find that, indeed, for the majority of the 100 instances, the input questions are semantically equal to the demonstration questions.

\begin{table}[ht]
    \centering
    \small
    \setlength{\tabcolsep}{5pt}
    \begin{tabular}{@{}ccccc@{}}
        \toprule
        \textbf{Task}  & \textbf{Random} & \textbf{Retriever} & \textbf{One-shot} & \textbf{Few-shot} \\
        \toprule
    \multirow{2}{*}{ANLI3} & \multirow{2}{*}{{33.33}}  
    & BM25 & 33.33 & 31.42 \\
    & & GTR & 34.75 & 32.25 \\
    \midrule
    \multirow{2}{*}{{StrategyQA }} & \multirow{2}{*}{{50.0}} 
    & BM25 & 48.79 & 47.34 \\
    & & GTR  & 47.83 & 48.31 \\
    \midrule
    \multirow{2}{*}{{AQUA }} & \multirow{2}{*}{{20.0 }} 
    & BM25 & 22.83 & 25.98 \\
    & & GTR & 24.02 & 22.05 \\
    \midrule
    \multirow{2}{*}{{GSM8K }} & \multirow{2}{*}{{0.0 }} 
    & BM25 & 1.36 & 1.82  \\
    & & GTR & 0.99 & 1.14 \\
    \midrule
    \multirow{2}{*}{{NQ }} & \multirow{2}{*}{{0.0 }} 
    & BM25 & 8.95& 8.70\\
    & & GTR & 11.99 & 11.08 \\
    \bottomrule
    \end{tabular}
    \caption{Overlapped Ratio of Demonstrations Answers with Targets: \textbf{Random} represents the probability of selecting the correct label if we select randomly from the space of possible labels.   
    }
    \label{tab:analysis_fewshot} 
\end{table}

%% file: conclusion.tex
\section{Discussion and Conclusion}
In this work, we first leverage two off-the-shelf retrievers to enhance ICL by searching query-oriented demonstrations.
Our experiments demonstrated that off-the-shelf retrievers are more effective than random demonstrations, with GTR generally retrieving more representative demonstrations than BM25. 
More importantly, our results with Flan-PaLM indicated that training data can be useful not only for training a model but also for improving the performance of fine-tuned LLM during testing via ICL.
Our experiments with CoT also suggests that integrating \dricl{} with advanced prompting techniques can further improve model's performance. 
Additionally, we trained a demonstration retriever that further improved the overall performance of off-the-shelf retrievers, with the most significant improvements observed in the one-shot scenario. One interesting future research challenge is retrieving demonstrations across tasks in situations where training data is not available.

%% file: tables/examples.tex
\begin{table*}[t]
    \centering 
    \small
    \setlength{\tabcolsep}{3pt}
    \begin{tabular}{@{}p{0.32\linewidth}p{0.32\linewidth}p{0.32\linewidth}@{}}
        \toprule 
        \textbf{Question} & \textbf{BM25 Demo} & \textbf{GTR Demo} \\
        \toprule
        Q: when does the new episodes of supernatural start? \newline A: October 12, 2017
        & 
        Q: when does the new episodes of ghost adventures start? \newline A: June 16, 2018
        & 
        Q: when does the next episode of supernatural come out? \newline A: April 5, 2018\\
        \midrule
        Kaj Birket-Smith (20 January 1893 – 28 October 1977) was a Danish philologist and anthropologist. He specialized in studying the habits and language of the Inuit and Eyak. He was a member of Knud Rasmussen's 1921 Thule expedition. In 1940, he became director of the Ethnographic Department of the National Museum of Denmark. 
        \newline question:  Kaj Birket-Smith would have been a ripe old age of 128 if he were still alive today. Is it true, false, or neither?
        \newline answer: false
        & 
        Kaj Birket-Smith (20 January 1893 – 28 October 1977) was a Danish philologist and anthropologist. He specialized in studying the habits and language of the Inuit and Eyak. He was a member of Knud Rasmussen's 1921 Thule expedition. In 1940, he became director of the Ethnographic Department of the National Museum of Denmark. 
        \newline question:  Kaj Birket-Smith was on the Thule expedition. Is it true, false, or neither?
        \newline answer: true
        & 
        Kaj Birket-Smith (20 January 1893 – 28 October 1977) was a Danish philologist and anthropologist. He specialized in studying the habits and language of the Inuit and Eyak. He was a member of Knud Rasmussen's 1921 Thule expedition. In 1940, he became director of the Ethnographic Department of the National Museum of Denmark. 
        \newline question:  Kaj Birket-Smith was a very educated man about many different cultures and expressed love in his field of expertise. Is it true, false, or neither?
        \newline answer: neither \\
        \midrule
        Q: The original retail price of an appliance was 60 percent more than its wholesale cost. If the appliance was actually sold for 20 percent less than the original retail price, then it was sold for what percent more than its wholesale cost?
        \newline Options:
        (A) 20%
        (B) 28%
        (C) 36%
        (D) 40%
        (E) 42%
        Step-by-step reasoning process: wholesale cost = 100;
        original price = 100*1.6 = 160;
        actual price = 160*0.8 = 128.
        \newline A: (B)
        & 
        Q: A retail appliance store priced a video recorder at 20 percent above the wholesale cost of \$200. If a store employee applied the 20 percent employee discount to the retail price to buy the recorder, how much did the employee pay for the recorder?
        \newline Options:
        (A)  \$198
        (B)  \$216
        (C)  \$192
        (D)  \$230
        (E)  \$240
        \newline  Step-by-step reasoning process: Wholesale cost of video recorder = 200 \$
        Video recorder was priced at 20 percent above 200 = 240 \$
        \% discount given by store employee = 20
        Emlpoyee paid = .8 * 240 = 192 \$ 
        \newline A: (C)
        & 
        Q: A retailer bought a machine at a wholesale price of \$108 and later on sold it after a 10\% discount of the retail price. If the retailer made a profit equivalent to 20\% of the whole price, what is the retail price of the machine?
        \newline Options:
        (A) 81
        (B) 100
        (C) 120
        (D) 135
        (E) 144
        \newline Step-by-step reasoning process: My solution: Wholesale Price= 108
        Retail Price, be = x
        He provides 10 \% discount on Retail price= x-10 x/100
        This Retail price = 20 \% profit on Wholesale price
        x-10 x/100 = 108+ 1/5(108)
        x=144;
        \newline A: (E)
        \\
        \midrule
        Q: Lori wants to buy a \$320.00 pair of shoes and a matching belt that is \$32.00.  Her part-time job pays her \$8.00 an hour.  How many hours will she have to work before she can make her purchase?
        \newline Step-by-step reasoning process: b"She wants to buy a pair of shoes for \$320.00 and a belt for \$32.00 for a total of 320+32 = \$<<320+32=352.00>>352.00
        Her purchase will total \$352.00 and she makes \$8.00 at her part-time job so she'll have to work 352/8 = <<352/8=44>>44 hours
        \newline A: 44
        & 
        Q: Joanne makes \$16.00 working at her main job for 8 hours a day. She has a part-time job, working an extra 2 hours a day where she makes \$13.50 an hour. How much money does she make if she works this schedule 5 days a week?
        \newline Step-by-step reasoning process: She works 8 hours a day at \$16.00 an hour so she makes 8 * 16 = \$128.00 a day. She works this job 5 days a week so she makes 128 * 5 = \$640.00 in 5 days. She works 2 hours a day at \$13.50 an hour so she makes 2 * 13.50 = \$27.00 a day. She works this job 5 days a week so she makes 27 * 5 = \$135.00. She makes \$640 at her main job and \$135 at her part - time job so all total she makes 640 + 135 = \$775.00 in 5 days.
        \newline A: 775
        & 
        Q: Janice has been working part-time at a convenience store 5 days a week. She can earn \$30 per day and can earn \$15 more when she works a 2 hour overtime shift. If she works three overtime shifts this week, how much will she earn this week?
        \newline Step-by-step reasoning process: Janice can earn \$30 x 5 = \$150 per week. She will earn \$15 x 3 = \$45 more if she works three overtime shifts. Therefore, Janice will earn \$150 + \$45 = \$195 this week.
        \newline A: 195
        \\
        \midrule
        Q: If it socially acceptable to wear an icon depicting crucifixion? 
        \newline Step-by-step reasoning process: The crucifixion of Jesus is a common sign used by Catholics and Christian denominations.Many jewelry stores offer necklaces with the Crucifixion of Jesus Christ.
        \newline A: yes
        &
        Q: Was the Donatello crucifix identified in 2020 life size?
        \newline Step-by-step reasoning process: The crucifix discovered in the church of Sant’Angelo depicts an adult man. The crucifix discovered in the church of Sant’Angelo is 89 cm high. The crucifix discovered in the church of Sant'Angelo was identified as being a work of Donatello. The average height of an adult man has been at least 150 cm in historical times.
        \newline A: no
        & 
        Q: Did any cultures associate celery with death?
        \newline Step-by-step reasoning process: Ancient Greeks used garlands of celery leafs to bury their dead. Ancient Greece was considered a culture.
        \newline A: yes \\
        \bottomrule
    \end{tabular}
    \caption{
    Examples of retrieved demonstrations.
    NQ, ANLI(r3), AQUQ, GSM8K, StrategyQA}
    \label{tab:example}
\end{table*}